\documentclass[twoside,11pt]{article}

\usepackage{blindtext}
\usepackage{arxiv}
\usepackage{amsmath}
\usepackage{algorithm}
\usepackage{algpseudocode}

\usepackage{lastpage}
\jmlrheading{}{2024 Fall}{1-\pageref{LastPage}}{6/30; Revised 11/10}{11/25}{21-0000}{Star (Xinxin) Liu}

\ShortHeadings{Machine Learning Paradigms via Statistical Thermodynamics: A tutorial}{ML tutorial via Stat Mech}
\firstpageno{1}

\begin{document}

\title{Understanding Machine Learning Paradigms through the Lens of Statistical Thermodynamics: A tutorial}

\author{\name Star (Xinxin) Liu \email starliu\@seas.upenn.edu \\
       \addr Department of Computer and Information Science \\
       Department of Materials Science and Engineering\\
       University of Pennsylvania\\
       Philadelphia, PA 19104, USA
       }

\editor{}
\maketitle

\begin{abstract}%   <- trailing '%' for backward compatibility of .sty file
This tutorial paper investigates the convergence of statistical mechanics and learning theory, elucidating the potential enhancements in machine learning methodologies through the integration of foundational principles from physics. The tutorial delves into advanced techniques like entropy, free energy, and variational inference which are utilized in machine learning, illustrating their significant contributions to model efficiency and robustness. By bridging these scientific disciplines, the tutorial aspires to inspire newer methodologies in researches, demonstrating how an in-depth comprehension of physical systems' behavior can yield more effective and dependable machine learning models, particularly in contexts characterized by uncertainty.
\end{abstract}


\section{Introduction}

Statistics is a classical discipline that deals with the general rules governing large groups of objects, where the impact from data errors and inaccuracies can be reduced through general modeling. Statistical mechanics provides an explanation for the behavior and movement of particles, applying statistical principles to physical systems. In the development of intelligent agents, various algorithms are designed using well-established theories, including Stat Mech. The intricate connections between statistical mechanics and learning theory offer unique insights for advancing both fields.\cite{carleo2019machine, watkin1993the}

At the root of both disciplines is the concept of probability distributions. In statistical mechanics, these distributions describe the likelihood of a system being in various energy states, providing insights into how systems reach equilibrium. In statistical learning, probability distributions model the data generation process, which is crucial for evaluating and improving machine learning models. The Central Limit Theorem, a fundamental result in probability theory, demonstrates how the distribution of the sum of a large number of independent random variables approximates a normal distribution. This theorem not only justifies the use of normal distributions for macroscopic properties in physics but also supports inference in machine learning by simplifying hypothesis testing and enhancing algorithm reliability. Another essential technique is importance sampling, which estimates properties of a target distribution by sampling from a different distribution. This method is critical in both thermodynamics for estimating properties in high-dimensional spaces and in machine learning for handling imbalanced data and optimizing training processes.\cite{watkin1993the}

By examining concepts such as entropy, free energy, and variational inference, this tutorial aims to elucidate how ideas from statistical mechanics can inform and enhance machine learning methodologies. This exploration seeks to provide a comprehensive understanding of how theoretical principles can be adapted to improve the efficiency and robustness of machine learning models, particularly in environments characterized by high uncertainty.

\section{Fundamental Theorem}

\textbf{Probability distributions} are essential for modeling and analyzing random phenomena in statistics. A probability distribution is a mathematical function that assigns probabilities to each possible outcome of a random variable, ensuring the total probability sums to one. For a discrete random variable $X$, the probability distribution is given by
\begin{equation}
    P(X = x_i) = p_i, \quad \text{where} \sum_i p_i = 1
\end{equation}
For a continuous random variable, the probability density function $f(x)$ satisfies
\begin{equation}
    \int_{-\infty}^{\infty} f(x) \, dx = 1
\end{equation}

In statistical mechanics, probability distributions describe the likelihood of a system being in various energy states, quantifying how system gradually reaches equilibrium. In statistical learning, probability distributions model the data generation process, describing the characteristics of data and in turn evaluating of machine learning models. 

\begin{flushleft}
The \textbf{Central Limit Theorem (CLT)} states that the distribution of the sum (or average) of a large number of independent, identically distributed random variables approaches a normal distribution, regardless of the original distribution of the variables. Briefly speaking, for independent and identically distributed (i.i.d.) random variables $ X_1, X_2, \ldots, X_n$ satisfy
\begin{equation}
    \forall i = 1, 2, \ldots n,\text{ follows } X_i \sim \mathcal{X}(\mu, \sigma^2), \quad \text{where }  \mu = \mathbb{E}[X_i] \text{ and } \sigma^2 = \text{Var}(X_i)
\end{equation}
\end{flushleft}
the sum $S_n = \sum_{i=1}^{n} X_i$ normalized by $\sqrt{n}$ converges to a normal distribution as $n$ increases.\vspace{3pt}

CLT justifies using normal distributions for macroscopic properties like temperature and pressure, derived from numerous microscopic interactions in statistical mechanics framework. Also, in theoretical machine learning, it supports inference by enabling normal distribution approximations for sample means, thus simplifying hypothesis testing, confidence intervals, enhancing ML algorithms and functionals' reliability and interpretability.

\begin{flushleft}
\textbf{Importance Sampling} is a statistical technique used to estimate properties of a target distribution by sampling from a different distribution. The key idea is to reweight samples from the proposal distribution to reflect their importance to the target distribution. Mathematically, if $p(x)$ is the target distribution and $q(x)$ is the proposal distribution, the expectation $\mathbb{E}_p[h(x)]$ can be estimated as:
\begin{equation}
    \mathbb{E}_p[h(x)] = \int h(x) p(x) \, dx = \int h(x) \frac{p(x)}{q(x)} q(x) \, dx \approx \frac{1}{N} \sum_{i=1}^{N} h(x_i) \frac{p(x_i)}{q(x_i)}
\end{equation}
\end{flushleft}
where $x_i$ are samples from $q(x)$.

Importance sampling is crucial for estimating thermodynamic properties by focusing on significant states, especially in high-dimensional integrals and rare-event systems. In ML algorithms, it is commonly used to handle imbalanced data and optimize training processes by emphasizing underrepresented yet highly influential samples, thereby enhancing the efficiency of training algorithms.

\subsection{Fundamentals in Statistical Mechanics }

A \textbf{micro-state} refers to the specific combinations of position $q_i$ and momentum $p_i$ of each particle $i$ in the system, corresponding to a phase point $Q(q,p)$ on a phase trajectory. To investigate the system's evolution over a given time and space interval, we use a set of statistical variables to describe their distribution, known as macro-state parameters. \textbf{Macro-state} parameters include extensive properties, which depend on the number of particles, and intensive properties, which are independent of the number of particles.

To further simplify the calculation, the \textbf{principle of equal a priori probabilities} states that micro-states with the same energy, volume, and number of particles occur with equal frequency in the ensemble. Thus, the probability density is written as:
\begin{equation}
\rho(q, p) = C \delta(H(q, p) - E)
\end{equation}
Intuitively, the number of micro-states of a system, $\Omega$, is a function of system energy $E$ and volume $V$. Ludwig Boltzmann introduced the \textbf{multiplicity} (or \textbf{Boltzmann principle} or \textbf{statistical weight}), which refers to the number of micro-states corresponding to a particular macro-state of a thermodynamic system:
\begin{equation}
S = k_B \ln \Omega \quad \text{or} \quad S(E, V) = k_B \ln \Omega(E, V)
\end{equation}
where $S$ is the entropy, $k_B = 1.38 \times 10^{-16}$ is Boltzmann's constant, and $\Omega$ is the number of micro-states corresponding to the macro-state. With a model to describe the possible micro-states of the system, we can calculate the thermodynamic properties of a bulk material.

\textbf{Thermal equilibrium} states that if system A is in thermal equilibrium with system B, and system B is in thermal equilibrium with system C, then system A is in thermal equilibrium with system C. \textbf{First Law of Thermodynamics} asserts that the energy of the universe (or any isolated system) is conserved and does not change over time. Any changes in energy, heat, or work within a system must balance out to result in zero net change in energy due to \textit{time-translation symmetry}. For any infinitesimal time span, the system's energy satisfies:
\begin{equation}
dE_{\text{sys}} = \delta W_{\text{on sys.}} + \delta Q_{\text{exit}} \quad \text{[Conservation of energy]}
\end{equation}
where the internal energy change of the system is $dE_{\text{sys}}$, external work done on the system is $\delta W_{\text{on sys.}}$, and heat leaving the system is $\delta Q_{\text{exit}}$. In a quasi-static process, which occurs slowly enough for the system to remain in internal thermodynamic equilibrium, we have $dW = -P dV$, so the First Law can be expressed as:
\begin{equation}
dE_{\text{sys}} = -P dV + \delta Q_{\text{exit}}
\end{equation}

\textbf{Second Law of Thermodynamics} states that heat cannot transfer from a colder to a warmer body without some other accompanying change. In any isolated system, entropy either increases or remains constant over time. When entropy remains constant, the system is in thermal equilibrium. The second law holds strictly for macroscopic systems but can be momentarily violated in microscopic systems due to thermal fluctuations. The \textbf{fluctuation theorem} quantitatively describes the probability of such entropy decreases. If the entropy change of a system is $\Delta S$, the probability $P(\Delta S)$ of this change compared to the probability $P(-\Delta S)$ of the opposite change is given by:
\begin{equation}
\frac{P(\Delta S)}{P(-\Delta S)} = e^{\Delta S / k_B}
\end{equation}
where $k_B$ is the Boltzmann constant. This relation shows that the probability of a decrease in entropy is significantly lower than the probability of an increase, and this probability decreases rapidly with system size.

In any \textbf{solid state} system (a system with a single micro-state in the ground state), the heat capacity approaches zero as the temperature approaches absolute zero:
\begin{equation}
\lim_{T \to 0} C_y = 0
\end{equation}
\textbf{Third Law of Thermodynamics} states that the entropy of a crystalline solid (solid state) approaches zero as the temperature approaches absolute zero:
\begin{equation}
\lim_{T \to 0} \Delta S_T = 0
\end{equation}
It is impossible to reach absolute zero temperature through successive finite implementations of a cyclic process.

\subsection{Fundamentals in Theoretical Machine Learning}

\textbf{Approximation algorithms} for NP-complete (NPC) problems usually run in polynomial time. In addition to the commonly discussed greedy and randomized algorithms, heuristic algorithms also provide a reasonable way to find approximate solutions with acceptable errors compared to the true optimal solution in a reasonable search time. Many ideas such as this have inspired research in the field of artificial intelligence and machine learning.

In a subset of NPC problems, optimization problems, the performance ratio of approximate algorithms discusses the relationship between the cost of the algorithm in solving the problem and the cost of the optimal solution. Specifically, let $C$ be the cost of an algorithm and $C^*$ be the cost of the optimal solution. The approximation ratio $ \rho(n)$ represents the difference factor between the cost $C$ of the solution generated by the algorithm and the cost $C^*$ of the optimal solution. For any input size $n$, the formula
\begin{equation}
\max\left(\frac{C}{C^*}, \frac{C^*}{C}\right) \leq \rho(n)
\end{equation}
shows that this ratio does not exceed a certain function $\rho(n)$. Algorithms that achieve an approximation ratio of $ \rho(n)$ are called $ \rho(n)$-approximation algorithms.

Machine learning (ML) addresses NP-complete problems by extracting patterns from data to provide  approximate solutions. Learning theory helps by understanding and designing algorithms that generalize well and perform reliably. It focuses on minimizing expected loss through empirical risk minimization.

We assume that the training data $\{(X_i, Y_i)\}_{i=1}^{n}$ are independently and identically distributed (i.i.d.) from an unknown, stationary distribution $\mathcal{D}$. The data points are assumed to be "sprinkle" within a finite sample space $\mathcal{X} \times \mathcal{Y}$. Furthermore, we assume that the observations include noise following a certain distribution.

Each ML model can be viewed as a family of functionals parameterized by hyper-parameters $\theta$, forming a \textbf{hypothesis space} $\mathcal{H}_{\theta}$. For a specific set of hyper-parameters $\theta$, the functional $F_{\theta}$ represents a set of parameterized functions $f_{\theta}(w)$, where $w$ denotes the parameters of the function.
Given the target function $f^*$ we aim to approximate, we use a loss function $L : \mathcal{Y} \times \mathcal{Y} \to \mathbb{R}^+$ to measure the difference between the predicted value $f_{\theta}(w; X)$ and the true value $Y$. The iterative process, often implemented via gradient descent or its variant methods, aims to find the parameters $w$ that minimize the expected loss:
\begin{equation}
R_{\theta}(w) = \mathbb{E}_{(X, Y) \sim \mathcal{D}} [L(f_{\theta}(w; X), Y)]
\end{equation}
Since the distribution $\mathcal{D}$ is unknown, we approximate the expected loss by minimizing the empirical risk:
\begin{equation}
\hat{w} = \arg \min_{w} \hat{R}_{\theta}(w) = \arg \min_{w} \frac{1}{n} \sum_{i=1}^{n} L(f_{\theta}(w; X_i), Y_i)
\end{equation}

\textbf{Probably Approximately Correct} (PAC) learning explores the theory for understanding the computational complexity of learning tasks. It quantifies the learnability of functions based on sample size and computational effort, providing a rigorous foundation for analyzing learning algorithms. In PAC learning, an algorithm seeks a hypothesis from a class that approximates the true function within given accuracy and confidence levels.

For any target function $f$ which represents the true concept
\begin{equation}
    f : \mathcal{X} \to \mathcal{Y}
\end{equation}
there is an approximation $h \in \mathcal{H}$ of $f$ with an error err$(h)$ satisfies that
\begin{equation}
    \text{err}(h) = \Pr_{(x,y) \sim D}[h(x) \neq y] \quad \leq \epsilon
\end{equation}

A class, or we say hypothesis space $\mathcal{H}$, is PAC-learnable only if there exists an algorithm $\mathcal{A}$ that, for any accuracy $\epsilon$ and confidence level $\delta$, and any distribution $D$ over $\mathcal{X} \times \mathcal{Y}$, outputs $h \in \mathcal{H}$ 
where the probability should be at least $1 - \delta$, given a sample size
\begin{equation}
    m \geq \frac{1}{\epsilon} \left( \log \frac{|\mathcal{H}|}{\delta} + k \right)
\end{equation}
where $k$ is a constant, and the algorithm $\mathcal{A}$ should be efficient, i.e. its running time is polynomial in $1/\epsilon$, $1/\delta$, and $m$. This framework has significantly impacted computational learning theory and AI by offering a structured method to evaluate and develop learning algorithms.\cite{valiant1984theory,valiant2013probably}

Based on PAC learnable framework, there are two definations. If we say an algorithm is \textbf{strongly learnable}, there is a concept class $C$ is strongly learnable if there exists an algorithm $A$ such that for any concept $c \in C$, any distribution $D$, and any $\epsilon > 0$, the algorithm $A$ outputs a hypothesis $h$ with:
\begin{equation}
    \Pr_{v \sim D}[h(v) = c(v)] \geq 1 - \epsilon
\end{equation}
If we say an algorithm is \textbf{weakly learnable}, there is a concept class $C$ is weakly learnable if there exists an algorithm $A$ and a constant $\gamma > 0$ such that for any concept $c \in C$ and any distribution $D$, the algorithm $A$ outputs a hypothesis $h$ with:
\begin{equation}
    \Pr_{v \sim D}[h(v) = c(v)] \geq \frac{1}{2} + \gamma
\end{equation}
\textbf{Hypothesis Boosting:} Start with distribution $D_1 = D$ and use the weak learner $A$ to get $h_1$:
\begin{equation}
\Pr_{v \sim D_1}[h_1(v) = c(v)] \geq \frac{1}{2} + \gamma
\end{equation}
\textbf{Iterative Distribution Modification:} \\
Step 1: Create distribution $D_2$ emphasizing instances mis-classified by $h_1$ and use $A$ to get $h_2$:
\begin{equation}
    \Pr_{v \sim D_2}[h_2(v) = c(v)] \geq \frac{1}{2} + \gamma
\end{equation}
Step 2: Create distribution $D_3$ based on mis-classifications of $h_1$ and $h_2$ to get $h_3$:
\begin{equation}
    \Pr_{v \sim D_3}[h_3(v) = c(v)] \geq \frac{1}{2} + \gamma
\end{equation}
\textbf{Combining Hypotheses:} Combine $h_1$, $h_2$, and $h_3$ using majority vote:
\begin{equation}
        h(v) = \text{MajVote}(h_1(v), h_2(v), h_3(v))
\end{equation}
This reduces the error rate:
\begin{equation}
        \text{Error}(h) \leq 3 \left( \frac{1}{2} - \gamma \right)^2 - 2 \left( \frac{1}{2} - \gamma \right)^3
\end{equation}
By recursively applying this process, the error can be reduced to any desired level $1 - \epsilon$. Thus, strong learnable and weak learnable are equivalent, Q.E.D.\cite{schapire1990strength}

\section{Adapting Ideas into Theoretical Machine Learning}

Many theoretical concepts from statistical mechanics are transferred to enhance machine learning methodologies. Key ideas include entropy, free energy, and variational inference. Entropy measures system disorder and information uncertainty, while free energy guides equilibrium states and informs variational inference to handle uncertainty in reinforcement learning. Advanced techniques also include like energy-based learning, the Ising model, and convolution. These methods improve model efficiency and robustness, particularly in uncertain environments.

\subsection{Entropy and Information Gain}

In the scenario of thermodynamics, for any micro-canonical ensemble (N, E, V are constant), the fundamental quantity Gibbs entropy is given by
\begin{equation}
S = -k_B \sum_i p_i \ln p_i = k_B \ln \Omega
\label{eq:gibbs_entropy}
\end{equation}
where the summation is over all possible micro-states of system, $\Omega$ is the "statistical weight" of micro-states in Boltzmann defination, and $p_i$ is the probability of a system being in micro-state $i$.

Transferred from such concept of system's disorder, Shannon gives a gauge describing uncertainty of information:
\begin{equation}
    I = - \sum_{i=0}^{W-1} p_i \log_2 p_i
    \label{eq:entropy}
\end{equation}
which is also called Information Gain (IG) in decision tree (DT) pruning.

A variant of Shannon Information used in message sequential is called algorithmic information. It is hard to express the $K$, the Shannon Information of one certain sequential.\cite{chaitin1975randomness,machta1999entropy} However, their ensemble property, average algorithmic information, can be expressed as:
\begin{equation}
    \langle K \rangle = \sum_i p_i K(s_i) \approx I
\end{equation}
where $I$ is Shannon information in Eq.~\ref{eq:entropy}.
\cite{machta1999entropy}

\subsection{Free Energy and Variational Inference} \label{section:free_energy_vari_inference}

In statistical mechanics, systems tend to maximize entropy, or disorder. However, for isothermal systems considering temperature and internal energy, Helmholtz Free Energy $A$ better describes equilibrium by accounting for both entropy and internal energy:
\begin{equation}
A = U - TS
\end{equation}
where $U$ is the internal energy, $T$ is the temperature, and $S$ is the entropy. This state function is used to determine the equilibrium state under the condition of free energy minimization.

In reinforcement learning (RL), solely maximizing rewards can be insufficient in complex environments. Friston introduces variational inference to handle uncertainty by minimizing free energy, which quantifies the difference between predicted and observed outcomes. Information gain (IG) or entropy, representing epistemic value, influences agent behaviors based on environmental perceptions. \cite{friston2015active, friston2016active} For updating beliefs by minimizing free energy efficiently, the variational free energy $F$ defined as
\begin{equation}
F = \mathbb{E}[\text{Model Energy} - \text{Entropy}] = D_{KL}(Q(Z) \parallel P(Z|X))
\end{equation}
where $D_{KL}$ is the Kullback-Leibler divergence, $Q(Z)$ is the approximate posterior distribution, and $P(Z|X)$ is the true posterior distribution given data $X$. Active inference uses variational free energy to guide the learning process. Agents select actions that minimize expected free energy, balancing exploration (epistemic value) and exploitation (extrinsic value). This is formalized as
\begin{equation}
G(\pi) = \mathbb{E}[r(o, s) - \ln Q(s' | s, \pi)]
\label{eq:friston_free_energy}
\end{equation}
where $G(\pi)$ is the expected free energy under policy $\pi$, $r(o, s)$ is the reward, and $Q(s' | s, \pi)$ is the state transition probability.\cite{friston2016active}

By transferring the concept of free energy, active learning frameworks can address uncertainty and optimize learning processes, drawing parallels between physical systems and learning algorithms. This integration enhances the efficiency and robustness of machine learning models, particularly in environments with high uncertainty.

Bellman's equation in reinforcement learning defines the optimal policy in a Markov decision process (MDP). It expresses the value function $V(s)$ and expected rewards $Q(\pi)$ as
\begin{align}
V(s) &= \max_a \left[ R(s, a) + \gamma \sum_{s'} P(s' | s, a) V(s') \right] \\
Q(\pi) &= \sum_{t=0}^{T} \mathbb{E}_{P(s' | s, a)} \left[ R(s, a) + \gamma V(s') \right]
\end{align}
where $R(s, a)$ is the reward for action $a$ in state $s$, $P(s' | s, a)$ is the transition probability to state $s'$, and $\gamma$ is the discount factor. Incorporating free energy Eq.~\ref{eq:friston_free_energy} into Bellman's equation, the value function and action-value function becomes
\begin{align}
V(s) &= \min_\pi \left[ \mathbb{E}[r(o, s) - \ln Q(s' | s, \pi)] + \gamma \sum_{s'} P(s' | s, \pi) V(s') \right] \\
Q(\pi) &= \sum_{t=0}^{T} \mathbb{E}_{Q(o_t, s_t | \pi)} \left[ \ln P(o_t, s_t | \pi) - \ln Q(s_t | \pi) \right]
\end{align}
where $P(o_t, s_t | \pi)$ is the generative model of observations and states given policy $\pi$, $Q(s_t | \pi)$ is the approximate posterior distribution over states given policy $\pi$.

Thus, while Bellman's original equation maximizes cumulative rewards, integrating free energy minimizes both expected free energy and uncertainty. This provides a unified approach, combining reinforcement learning's reward optimization with active inference's uncertainty reduction. \cite{friston2015active}

\subsection{Energy-Based Learning}

Energy-Based Learning (EBL) is a machine learning framework that models dependencies between variables by associating a scalar energy $E(Y, X)$ with each configuration of the variables $Y$ and $X$. The goal is to make correct configurations have lower energy than incorrect ones.\cite{lecun2006tutorial} \\
\textbf{Inference} finds the configuration $Y$ minimizes the energy for given observed variables $X$ as
\begin{equation}
Y^* = \arg \min_Y E(Y, X)
\end{equation}
\textbf{Learning} process adjusts the parameters $W$ of the energy function $E(Y, X; W)$ so that correct configurations from the training data have lower energies than incorrect ones.

EBL integrates concepts from statistical mechanics. The energy function indicates compatibility of variable configurations, where lower energy means higher compatibility. The Gibbs distribution provides a probabilistic interpretation:
\begin{equation}
P(Y | X) = \frac{e^{-\beta E(Y, X)}}{\sum_{y \in \mathcal{Y}} e^{-\beta E(y, X)}}
\end{equation}
with $\beta$ controlling the distribution's sharpness, and the partition function $Z(X)$ ensuring proper normalization:
\begin{equation}
Z(X) = \sum_{y \in \mathcal{Y}} e^{-\beta E(y, X)}
\end{equation}
\textbf{Perceptron Loss} focuses on the difference between the energy of the correct configuration and the lowest energy of any configuration:
\begin{equation}
L_{\text{perceptron}} = E(Y_i, X_i) - \min_Y E(Y, X_i)
\end{equation}
\textbf{Hinge Loss} introduces a margin $m$ to create a gap between the energy of correct and incorrect configurations:
\begin{equation}
L_{\text{hinge}} = \max(0, m + E(Y_i, X_i) - E(Y_{\text{incorrect}}, X_i))
\end{equation}
\textbf{Negative Log-Likelihood Loss} combines the energy of the correct configuration with the log of the partition function:
\begin{equation}
L_{\text{nll}} = E(Y_i, X_i) + \frac{1}{\beta} \log \left( \sum_{y \in \mathcal{Y}} e^{-\beta E(y, X_i)} \right)
\end{equation}

\subsection{Ising Model and Boltzmann Machine}

The Ising model is a mathematical model used to describe magnetic systems, where each particle can be in one of two spin states (+1 or -1). The energy of the system is determined by the interactions between neighboring particles, typically expressed as:
\begin{equation}
    E = - \sum_{i,j} J_{ij} s_i s_j - \sum_i h_i s_i
\end{equation}
where $s_i$ and $s_j$ are the spin states of particles $i$ and $j$, $J_{ij}$ is the interaction strength between them, and $h_i$ represents the external magnetic field. The partition function is expressed as
\begin{equation}
    Z = \sum_{\{s\}} e^{-E(s)}
\end{equation}
where $s$ represents all possible spin configurations.

Inspired by Ising model, Boltzmann machines to describe the energy and state probabilities of neural networks. The energy function for a Boltzmann machine is given by:
\begin{equation}
    E(\vec{v}, \vec{h}) = - \sum_{i \in \text{visible}} a_i v_i - \sum_{j \in \text{hidden}} b_j h_j - \sum_{i,j} v_i h_j w_{ij}
\end{equation}
where $v_i$ and $h_j$ are the states of visible and hidden units, $a_i$ and $b_j$ are their biases, and $w_{ij}$ are the weights connecting them. The normalization function is written as
\begin{equation}
     Z = \sum_{\vec{v}, \vec{h}} e^{-E(\vec{v}, \vec{h})}
\end{equation}
where $\vec{v}$ and $\vec{h}$ represent all possible visible and hidden unit configurations in the form of vectors. \cite{ackley1985a, hinton2012practical}

\subsection{Convolution and Smoothing}

For two independent random variables $X$ and $Y$ with distributions $f_X(x)$ and $f_Y(y)$, their sum $Z = X + Y$ has a distribution:
\begin{equation}
f_Z(z) = f_X * f_Y = \int_{-\infty}^{\infty} f_X(x) f_Y(z - x) \, dx
\end{equation}
Thus, convolution can describe the combined state of two independent systems $A$ and $B$ with probability distributions $p_A(x)$ and $p_B(y)$. The combined system $C$ has a distribution $p_C(z)$ given by:
\begin{equation}
p_C(z) = p_A * p_B = \sum p_A(x) p_B(z - x)
\end{equation}
In signal processing, the signal $f(t)$ with a kernel $h(t)$, the smoothed signal $g(t)$ is obtained via convolution:
\begin{equation}
g(t) = f(t) * h(t) = \int_{-\infty}^{\infty} f(\tau) h(t - \tau) \, d\tau
\end{equation}
For the matrix of input signal, vector $a=(a_0, a_1, \ldots, a_{n-1})$ and $b=(b_0, b_1, \ldots, b_{n-1})$
\begin{align}
    A(x) &= a_0 + a_1 x + a_2 x^2 + \ldots + a_{n-1} x^{n-1} \\
    B(x) &= b_0 + b_1 x + b_2 x^2 + \ldots + b_{n-1} x^{n-1} \\
\text{Then, } C(x) &= A(x) B(x) \\
        &= a_0 b_0 + (a_0 b_1 + a_1 b_0) x + \ldots + a_{n-1} b_{n-1} x^{2n-2}
\end{align}
Obviously, the coefficient vector of $C(x)$ is $a * b$. The convolution $a * b$ calculation is equivalent to multiplying polynomials. Convolution operation takes $O(n^2)$ runtime. Here, we introduce \textbf{Fast Fourier Transform} (FFT) Algo.~\ref{FFT-psuedocode}

\begin{algorithm}
\caption{Fast Fourier Transformation}
\begin{algorithmic}[1]
\For{$j = 0$ to $2n-1$}  \hfill $O(n \log n)$
    \State Evaluate $A(\omega_j)$ and $B(\omega_j)$
\EndFor
\For{$j = 0$ to $2n-1$}  \hfill $O(n)$
    \State Calculate $C(\omega_j)$
\EndFor
\State Construct polynomial:  \hfill $O(n \log n)$
\begin{equation*}
    D(x) = C(\omega_0) + C(\omega_1) x + \ldots + C(\omega_{2n-1}) x^{2n-1}
\end{equation*}
\For{$j = 0$ to $2n-1$}  \hfill $O(n)$
    \State Calculate $D(\omega_j)$
\EndFor
\State Calculate the coefficients $c_j$ of $C(x)$:  \hfill $O(n \log n)$
\begin{align*}
D(\omega_j) &= 2n c_{2n-j}, \quad j = 1, \ldots, 2n-1 \\
D(\omega_0) &= 2n c_0
\end{align*}
\end{algorithmic} \label{FFT-psuedocode}
\end{algorithm}
Then, the overal runtime complexity of convolution operation by FFT is $O(n \log n)$.

\subsection{Information Bottleneck Theory}

Information Bottleneck (IB) principle further discussed the trade-off between energy (relevance, $I(T; Y)$) and entropy (zipping complexity, $I(X; T)$), analyzing and optimizing DNNs. They propose that DNNs can be quantified by mutual information $I(X; T)$ and $I(T; Y)$ between input $X$, output $Y$, and hidden layers $T$. The goal is to find a compressed representation $T$ that retains maximal relevant information about $Y$, formulated as minimizing $L[p(t|x)] = I(X; T) - \beta I(T; Y)$, balancing complexity and predictive power. DNNs form a Markov chain
\begin{equation}
    X \to T_1 \to T_2 \to \cdots \to T_L \to Y
\end{equation}
where each layer $T_i$ refines information progressively. The Data Processing Inequality ensures information loss in one layer is irreversible, implying layers should maximize $I(T_{i-1}; T_i)$ while minimizing $I(T_i; Y)$.\cite{tishby2015deep}

The IB principle offers new finite sample complexity bounds on DNN generalization, guiding the optimal design of layers for better generalization and efficiency. This framework aligns the optimal DNN structure with the IB tradeoff curve, informing the development of more effective DL algorithms.

\subsection{Other Optimizors and Minimization Techniques}

Molecular Dynamics (MD) simulation is one of the most important computational application from statistical mechanics. MD and ML models both use iterative optimization processes but have different frameworks and objectives.

In MD simulations, systems evolve through discrete time steps, updating particle positions and velocities based on Newton's equations of motion. The goal is to minimize the system's potential energy, reaching equilibrium states where the energy function's partial derivatives are zero. Methods like conjugate gradient or L-BFGS are used, with computational demands scaling with the number of particles and interaction complexity.

In machine learning, models are trained iteratively using algorithms like stochastic gradient descent or Adam to minimize an expert-defined loss function. The objective is to find the optimal model parameters, where the gradient of the loss function is zero, similar to finding equilibrium in energy minimization. Convergence is assessed using performance metrics on validation or test datasets. Despite their different contexts, both MD simulations and machine learning aim to iteratively optimize a target function to achieve minimal energy or error.

More cutting edge researches focus on optimization techniques. The Replica method, developed by Giorgio Parisi, computes replicated partition sums and applies the Replica Symmetry Ansatz to solve high-dimensional optimization problems.\cite{mezard1985replicas,abbaras2020rademacher,mezard1987spin} The Cavity method, pioneered by  Marc Mézard, Giorgio Parisi and Miguel Angel Virasoro, derives self-consistent equations for large-scale random systems and is useful in studying phase transitions.\cite{mezard1987spin} Approximate Message Passing (AMP), introduced by David Donoho and Andrea Montanari, handles large-scale inference problems such as random matrices, focusing on convergence properties and applications in sparse vector denoising, low-rank matrix factorization, and community detection.\cite{donoho2009message,rangan2019vector} These methods, rooted in the statistical physics of spin glasses, provide powerful tools for addressing complex problems in modern computational science.

\subsection{Mean-Field Theory}

In statistical mechanics, real-world interactions are inherently multi-body problems where each particle influences all others. To simplify such complex systems, mean field theory (MFT) assumes that each particle is affected by an average field generated by all other particles. Mathematically, if we denote the state of particle $i$ by $\sigma_i$ and the interaction energy between particles $i$ and $j$ by $J_{ij} \sigma_i \sigma_j$, MFT approximates the interaction term as 
\begin{equation}
    \sigma_i \langle \sum_j J_{ij} \sigma_j \rangle
\end{equation}
where $\langle \cdot \rangle$ denotes the average. This reduces the complexity by transforming the multi-body problem into a single-body problem.

Building on this concept, mean field variational inference in machine learning simplifies the estimation of posterior distributions in probabilistic models. For a model with hidden variables $h$ and observed data $x$, the goal is to approximate the posterior $ p( h|x)$. MFT assumes a factorized form 
\begin{equation}
    q(h) = \prod_i q(h_i)
\end{equation}
and optimizes the variational parameters by minimizing the Kullback-Leibler divergence
\begin{equation}
    D_{KL}(q(h) \parallel p(h|x))
\end{equation}
This approach reduces the computational complexity from dealing with the full joint distribution to handling individual distributions.

Extending this idea further, deep learning neural networks (NNs) abstract and simulate the behavior of neurons in the human brain. For a single layer in a classical neural network, with input $x$, weights $W$, and activation function $\phi$, the output $y$ is given by $ y = \phi(W x)$. Thus, the effects of interactions between neurons are typically ignored, simplifying complicate communications into integral and homogeneous property. This mirrors the principles of mean field theory, where the complex interactions within the system are approximated by an average effect, allowing for efficient training and analysis of neural networks.

\subsection{From Markov Chain Monte Carlo to Mean Field Multi-Agent Reinforcement Learning}

Monte-Carlo algorithms are widely used, and the Metropolis algorithm's original paper is regarded as a landmark in computational physics. Here we show a general set of steps in Algo.~\ref{metroplis}.
\begin{algorithm}
\caption{Metropolis Algo for symmetric $q(s'|s)$}
\begin{algorithmic}
\State Initialise $s = s_0$;
\For{$t = 1, \ldots, T$}
    \State Suggest a new state $s'$ with probability $q(s'|s)$;
    \State Compute $\Delta H = H(s') - H(s)$;
    \If{$\Delta H \leq 0$}
        \State Accept the new state: $s = s'$;
    \Else
        \State Draw a random number $r$ uniformly distributed in $[0, 1]$;
        \If{$r < \exp(-\beta \Delta H)$}
            \State Accept the new state: $s = s'$;
        \Else
            \State Reject $s'$;
        \EndIf
    \EndIf
    \State Sample $s_t = s$ and $A_t = A(s_t)$;
\EndFor
\end{algorithmic} \label{metroplis}
\end{algorithm}

Markov Chain Monte Carlo (MCMC) methods, originally used in statistical and computational thermodynamics to estimate complex probability distributions, efficiently handle high-dimensional, multi-modal distributions by constructing a converging Markov chain. This approach was adapted to reinforcement learning (RL) for optimizing policies in complex state spaces to maximize cumulative rewards.

RL involves agents learning through iterative interaction and feedback from the environment. Agents observe the state $s_t$, take actions $a_t$ , and receive rewards $r_{t+1}$, aiming to learn an optimal policy $\pi$ to maximize cumulative rewards. \cite{liu2024recent} As mentioned in section~\ref{section:free_energy_vari_inference}, classical Q-learning algorithms learn the Q-function to quantify expected rewards for actions in each state. Deep reinforcement learning (DRL) integrates deep neural networks with RL, leading to methods like deep Q-networks (DQNs) and policy gradient methods.

Further RL advances includes Mean Field Multi-Agent Reinforcement Learning (MF-MARL) addresses scalability issues in traditional multi-agent RL (MARL) methods, which struggle with the curse of dimensionality. MF-MARL approximates interactions within a population by considering the average effect from neighboring agents. This simplifies multi-agent interaction into learning the optimal policy for an individual agent. \cite{yang2018mean} Mathematically, for an agent $j$ with neighboring agents $\mathcal{N}(j)$, the Q-function is approximated by applying the mean field theory:
\begin{equation}
    Q^j(s, a^j, a^{\mathcal{N}(j)}) \approx Q^j(s, a^j, \bar{a}^{\mathcal{N}(j)})
\end{equation}
where $\bar{a}^{\mathcal{N}(j)}$ is the mean action of the neighbors:
\begin{equation}
    \bar{a}^{\mathcal{N}(j)} = \frac{1}{|\mathcal{N}(j)|} \sum_{k \in \mathcal{N}(j)} a^k
\end{equation}
Then, the mean field Q-learning update rule is given by:
\begin{equation}
    Q^j_{t+1}(s, a^j, \bar{a}^{\mathcal{N}(j)}) = (1 - \alpha) Q^j_{t}(s, a^j, \bar{a}^{\mathcal{N}(j)}) + \alpha \left[ r^j + \gamma \mathbb{E}_{s'} \left[ V^j_{t}(s') \right] \right]
\end{equation}
where $\alpha$ is the learning rate, $r^j$ is the reward for agent $j$, and $V^j_{t}(s')$ is the value function at state $s'$. Similarly, in the mean field Actor-Critic method, the policy $\pi^j$ is updated using the gradient of the expected return:
\begin{equation}
    \nabla_{\theta_j} J(\theta_j) \approx \mathbb{E}_{s, a^j, \bar{a}^{\mathcal{N}(j)}} \left[ \nabla_{\theta_j} \log \pi_{\theta_j}(a^j | s) Q^j(s, a^j, \bar{a}^{\mathcal{N}(j)}) \right]
\end{equation}

Theoretical analysis proved that under certain conditions, the mean field Q-learning and Actor-Critic algorithms converge to a Nash equilibrium. Experiments on scenarios such as Gaussian squeeze, the Ising model, and battle games demonstrate the effectiveness and scalability of MF-MARL, solving complex multi-agent problems where traditional methods fail.\cite{yang2018mean}

\subsection{Simulated Annealing Searching}

In simulated annealing, we aim to minimize an energy function $H(s)$ to find the optimal configuration $s_{\text{min}}$. The process uses Monte-Carlo simulation to explore the energy landscape with stochastic dynamics, starting at a high temperature $k_B T = \beta^{-1}$ and gradually lowering it to refine the search. The Metropolis algorithm governs the acceptance of new configurations based on the probability \cite{kirkpatrick1983optimization}
\begin{align}
    P_B(s) &= 0 \quad \text{if } H(s) > H_{\text{min}} \\
    P_B(s) &> 0 \quad \text{if } H(s) = H_{\text{min}}
\end{align} 

As an example, in the double-digest problem, \cite{lander2001initial} enzymes $A$ and $B$ produce fragments $a = \{a_1, \ldots, a_n\}$ and $b = \{b_1, \ldots, b_m\}$. The goal is to determine the orderings of these fragments resulting in $c = \{c_1, \ldots, c_l\}$. The energy function for this problem is given by
\begin{equation}
    H(\sigma, \mu) = \sum_j c_j^{-1} [c_j - \hat{c}(\sigma, \mu)]^2
\end{equation}
where $\sigma$ and $\mu$ are permutations of $a$ and $b$, respectively. Stochastic dynamics, beginning at high temperature $k_B T = \beta^{-1}$ and gradually reducing it, are employed to minimize $H$ and explore the global minimum $H_{\text{min}}$. This method ensures that the dynamics effectively navigate both the rough and fine features of the energy landscape to find the optimal configuration.\cite{kirkpatrick1983optimization, lander2001initial}

Recent research applied SA gives a general optimizer for large AI model training. In recent bio-informatic works, a novel model named AbNatiV uses a Vector Quantized Variational Autoencoder (VQ-VAE) to measure how similar antibody sequences are to natural human or camel antibodies. It uses SA-based enhanced sampling to quickly identify promising mutants during training, helping it achieves state-of-the-art performance and shows potential in immunogenicity assessment.\cite{heydaribeni2024distributed}

\subsection{Sparse Ising Machine helps training Deep Boltzmann networks}

Researchers also investigate the utilization of sparse Ising machines (SIMs) for training deep Boltzmann networks (DBMs), presenting a substantial improvement in training efficiency and performance through specialized hardware architectures. The core concept leverages the Ising model to address combinatorial optimization problems using probabilistic-bit (p-bit) based Ising machines for efficient probabilistic sampling. Key techniques include the deployment of sparse, hardware-aware network topologies such as Pegasus and Zephyr graphs, enabling parallel and asynchronous updates that enhance sampling speed. \cite{niazi2024training}

Their results show that the SIM effectively trains the full MNIST and Fashion MNIST datasets without downsampling, achieving 90\% classification accuracy on MNIST with far fewer parameters than traditional restricted Boltzmann machines (RBMs). Additionally, the sparse DBM can generate new handwritten digits and fashion product images, which RBMs cannot do. It greatly improves training efficiency and hardware use especially in training deep generative AI models, especially in settings with limited resources. \cite{niazi2024training}

\section{Discussion}

Recent advancements in science-based artificial intelligence (AI) include the development of specialized models for prediction and the enhancement of computational science tools.

Modeling and forecasting the dynamics of multiphysics and multiscale systems, such as the Earth's system, present significant challenges due to the interactions across extensive spatial and temporal scales. These complexities render traditional methods computationally prohibitive. The integration of observational data is vital, as data-driven approaches improve the accuracy and feasibility of simulations. By combining machine learning with traditional numerical methods and embedding physical laws into models, more accurate and consistent results are achieved. Examples of such hybrid models include Physics-Informed Neural Networks (PINNs), equivariant neural networks and Network-of-Networks Model, which reduce the limitations of purely data-driven or physics-based approaches. \cite{karniadakis2021physics, pete2024physical}

AI has demonstrated its potential in solving complex problems such as protein folding and molecular simulations, underscoring its transformative power in scientific research. However, the application of AI in research is not without challenges. These include the substantial computational resources required, the need for data standardization, and the imperative of ensuring reproducibility of results. To enhance the effectiveness and reliability of AI models, interdisciplinary collaboration and the incorporation of domain-specific scientific knowledge are essential.

\section{Conclusion}

In this tutorial, we collect how statistical mechanics principles and discoveries inspire and enhance machine learning methodologies. By examining key concepts such as entropy, free energy, and variational inference, it has been demonstrated that these ideas can improve the efficiency, robustness, and interpretability of machine learning models. The basic intersection of these field also includes the Central Limit Theorem and importance sampling, provides a strong theoretical foundation for developing more reliable algorithms. Additionally, advanced techniques like mean field theory and simulated annealing have been shown to address complex optimization problems in machine learning effectively.

This synthesis of statistical mechanics and machine learning not only highlights the potential for interdisciplinary approaches to solve contemporary computational challenges but also sets the stage for future research that leverages the deep connections between these fields. The integration of these domains promises to yield innovative solutions and enhance our understanding of both physical systems and data-driven models in uncertain environments.

\newpage
\acks{
This work was supported by myself. 
We sincerely thank Professor Mahadevan Khantha from the University of Pennsylvania for constructive advices and technical support of our manuscript. \\
There are no conflicts to declare.
}


\newpage

\bibliography{reference}

\end{document}